\documentclass[10pt,twocolumn,letterpaper]{article}

\usepackage{wacv}
\usepackage{times}
\usepackage{epsfig}
\usepackage{graphicx}
\usepackage{amsmath}
\usepackage{amssymb}
\usepackage{xcolor}
\usepackage{pifont}

\newcommand{\cmark}{\ding{51}}
\newcommand{\xmark}{\ding{55}}
\renewcommand\fbox{\fcolorbox{purple}{white}}
\DeclareMathOperator{\E}{\mathbb{E}}

\newcommand{\algoname}{\textit{Disassembled Hourglass}}

\wacvfinalcopy % *** Uncomment this line for the final submission
\usepackage[breaklinks=true,bookmarks=false]{hyperref}
\usepackage{booktabs}
\usepackage[capitalise]{cleveref}

\Crefname{table}{Tab.}{Figs.}
\Crefname{section}{Sec.}{Figs.}

\begin{document}

%%%%%%%%% TITLE
\title{Unsupervised Novel View Synthesis from a Single Image}

\author{Pierluigi Zama Ramirez$^{1,2}$ \\
{\tt\small pierluigi.zama@unibo.it}\\\\\\\\\\
{$^1$University of Bologna}
\and
Diego Martin Arroyo$^2$ \\
{\tt\small martinarroyo@google.com} \\\\
Federico Tombari$^{2,3}$\\
{\tt\small tombari@google.com} \\\\
{$^2$Google Inc.}
\and
Alessio Tonioni$^2$ \\
{\tt\small alessiot@google.com} \\\\\\\\\\
{$^3$ Technische Universität München}
}

\maketitle

\begin{figure*}
	\setlength{\tabcolsep}{1pt}
	\setlength{\fboxrule}{2pt} 
	\setlength{\fboxsep}{0pt} 
	\centering
	\scalebox{0.98}{
	\begin{tabular}{cc|ccccccc}
	& \textit{Input.} & \multicolumn{7}{c}{\textit{Generated Views}} \\
	\rotatebox{90}{\hspace{1.5em} \small \textit{CelebA} }&\includegraphics[width=0.115\textwidth]{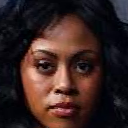} &
	\includegraphics[width=0.115\textwidth]{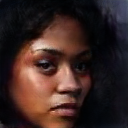} &
	\includegraphics[width=0.115\textwidth]{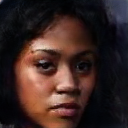} &
	\includegraphics[width=0.115\textwidth]{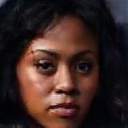} &
	\fbox{\includegraphics[width=0.115\textwidth]{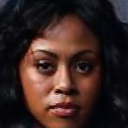}} &
	\includegraphics[width=0.115\textwidth]{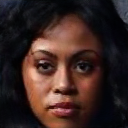} &
	\includegraphics[width=0.115\textwidth]{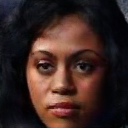} &
	\includegraphics[width=0.115\textwidth]{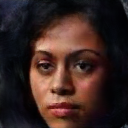} \\
	\rotatebox{90}{\hspace{1em} \small \textit{Real Cars}}&\includegraphics[width=0.115\textwidth]{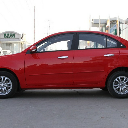} &
	\fbox{\includegraphics[width=0.115\textwidth]{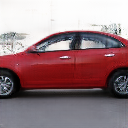}} &
	\includegraphics[width=0.115\textwidth]{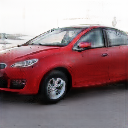} &
	\includegraphics[width=0.115\textwidth]{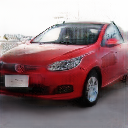} &
	\includegraphics[width=0.115\textwidth]{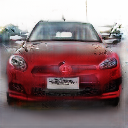} &
	\includegraphics[width=0.115\textwidth]{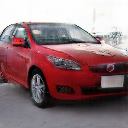} &
	\includegraphics[width=0.115\textwidth]{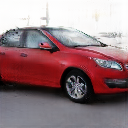} &
	\includegraphics[width=0.115\textwidth]{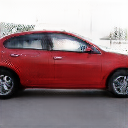} \\
	\rotatebox{90}{\hspace{0.1em} \small  \textit{ShapeNet Cars}}&\includegraphics[width=0.115\textwidth]{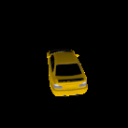} &
	\includegraphics[width=0.115\textwidth]{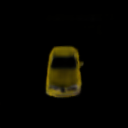} &
	\includegraphics[width=0.115\textwidth]{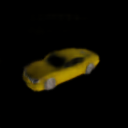} &
	\includegraphics[width=0.115\textwidth]{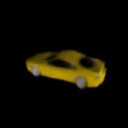} &
	\fbox{\includegraphics[width=0.115\textwidth]{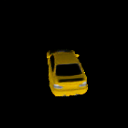}} &
	\includegraphics[width=0.115\textwidth]{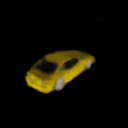} &
	\includegraphics[width=0.115\textwidth]{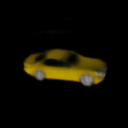} &
	\includegraphics[width=0.115\textwidth]{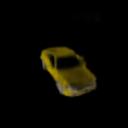} \\
	\rotatebox{90}{\hspace{0.1em} \small \textit{ShapeNet Sofa}}&\includegraphics[width=0.115\textwidth]{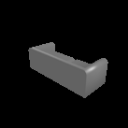} &
	\includegraphics[width=0.115\textwidth]{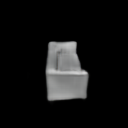} &
	\includegraphics[width=0.115\textwidth]{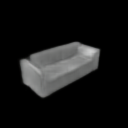} &
	\includegraphics[width=0.115\textwidth]{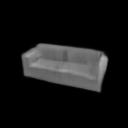} &
	\includegraphics[width=0.115\textwidth]{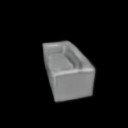} &
	\includegraphics[width=0.115\textwidth]{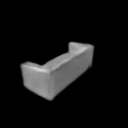} &
	\includegraphics[width=0.115\textwidth]{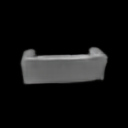} &
	\fbox{\includegraphics[width=0.115\textwidth]{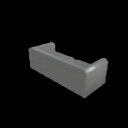}} \\

	\end{tabular}}
	\caption{Novel view synthesis from a single input image. We train our method on a set of natural images of a certain category (\eg faces, cars\dots) without any 3D, multi-view or pose supervision. 
	Given an input image (left column) our model estimates a global latent representation of the object and its pose to reconstruct the input image (images framed in red in each row). By changing the pose we can synthesize novel views of the same object (columns 2-8). }
	\label{fig:teaser}
\end{figure*}

\begin{abstract}
    Novel view synthesis from a single image has recently achieved remarkable results, although the requirement of some form of 3D, pose, or multi-view supervision at training time limits the deployment in real scenarios.
    This work aims at relaxing these assumptions enabling training of conditional generative models for novel view synthesis in a completely unsupervised manner.
    We first pre-train a purely generative decoder model using a 3D-aware GAN formulation while at the same time train an encoder network to invert the mapping from latent space to images.
    Then, we swap encoder and decoder and train the network as a conditioned GAN with a mixture of an autoencoder-like objective and self-distillation.
    At test time, given a view of an object, our model first embeds the image content in a latent code and regresses its pose, then generates novel views of it by keeping the code fixed and varying the pose.
    We test our framework on both synthetic datasets such as ShapeNet and on unconstrained collections of natural images, where no competing methods can be trained.
\end{abstract}

%\linenumbers

\section{Introduction}

Novel View Synthesis (NVS) aims to generate novel viewpoints of an object or a scene given only one or few images of it.
Given its general formulation, it is one of the most challenging and well-studied problems in computer vision, with several applications ranging from robotics, image editing, and animation to 3D immersive experiences.

Traditional solutions are based on multi-view reconstruction using a geometric formulation ~\cite{debevec1996modeling,fitzgibbon2005image,seitz2006comparison}.
Given several views of an object, these methods build an explicit 3D representation (\eg{} mesh, pointcloud, voxels\dots) and render novel views of it. 
A more challenging problem is novel view synthesis given just a single view of the object as input.
In this case, multi-view geometry cannot be leveraged, making this an intrinsically ill-posed problem.
Deep learning, however, has made it approachable by relying on inductive biases, similarly to what humans do.
For instance, when provided with the image of a car, we can picture how it would look from a different viewpoint.
We can do it because, unconsciously, we have learnt an implicit model for the 3D structure of a car, therefore we can easily \textit{fit} the current observation to our mental model and hallucinate a novel view of the car from a different angle.
This work tries to replicate a similar behaviour using a deep generative model.

Several works have recently investigated this line of research ~\cite{xu2019view,liu2020auto3d,olszewski2019transformable} by designing category-specific models that, at test time, can render novel views of an object given a single input view. 
However, all these methods require expensive information at training time such as 3D supervision ~\cite{rematasICML21} or multi-view images of the same object with known camera poses ~\cite{dupont2020equivariant, yu2021cvpr, liu2020auto3d, xu2019view, haeni2020corn}.
Our initial research question was: \textit{How can we learn to hallucinate novel views of a certain object from natural images only?}
Finding an answer would allow deployment to any object category without being constrained to synthetic benchmarks or well-crafted and annotation-intensive datasets.

Inspiration for our work comes from the recent introduction of 3D-aware GAN models ~\cite{Nguyen-Phuoc_2019_ICCV, nguyenphuoc2020blockgan, Liao_2020_CVPR, schwarz2020graf}, which allow to disentangle the content of a scene from the viewpoint from which it is observed.
These methods can \emph{generate} objects and synthesize views of them from a random latent vector. However, they cannot be directly used to synthesize novel views of a \emph{real} object unless some form of inversion or conditioning is applied.

We propose a novel framework for NVS based on inverting the knowledge of a 3D-aware GAN by means of self-distillation.
Our framework is composed of an encoder network that projects an input image to a latent vector and an estimated pose, and a decoder architecture that, from the estimated latent vector, recreates the input view (with the estimated pose) or generates novel views (by changing the pose).
We design a two-stage training procedure for this system.
In the first stage, we train the decoder as a 3D generative model, i.e., to synthesize images of new objects by sampling random latent representations. % from a random distribution.
At the same time we also pre-train the encoder to invert it by estimating the latent vector and pose of a generated image.
Then, in the second stage, we swap the position of encoder and decoder turning the network into a 3D autoencoder for NVS and train it to reconstruct real images.
Given the change of order between encoder and decoder in the two stages of our approach we named it \algoname{}.

Contrary to pre-existing GAN inversion methods ~\cite{zhuECCV2016, creswellTNNL2019} we propose to finetune the weights of the generator during this stage to achieve better reconstruction results.
In order to avoid a degenerate solution that does not maintain the 3D structure of the object, we propose to self-distill the knowledge of the generative model obtained in the first step.
In practice, we use it to generate multiple views for a sampled latent code, then we feed to the decoder one of the generated viewpoints and directly supervise it to reconstruct a different view of the same object. 
Supported by experimental results, we argue that this is key to learn a network able to achieve meaningful high quality NVS.
Finally, an optional third step finetunes the network on a specific image to recover fine details with just few optimization steps.

We show in \cref{fig:teaser} qualitative results of our framework on the standard benchmark for NVS ShapeNet ~\cite{shapenet2015}, as well as datasets of real images ~\cite{liu2015faceattributes,yang2015large}, where competitors cannot be trained due to the lack of pose annotations. 
We will release the code in case of acceptance.
To summarize our contributions:
\begin{itemize}
    \item We introduce a novel framework for NVS to generate novel views of a given object from a single input image, while at the same time regressing the image pose.
    
    \item We propose the first framework which can be trained on natural images without any form of explicit supervision besides representing objects of a certain category.
    
    \item We introduce a two-stage training approach that allows to self-distill the knowledge of a 3D-aware generative model to make it conditioned while preserving disentanglement of pose and content.
\end{itemize}

\begin{table}[t]
\centering
\scalebox{0.9}{
\centering
  \begin{tabular}{c|ccc|c}
     & \multicolumn{3}{c|}{Train} & Test \\
    \toprule
    Method & 3D & Pose & Multiview & Source Pose \\
    
    \midrule
    MV3D~\cite{tatarchenko2016multi} & \cmark & \cmark & \cmark & \xmark \\
    
    AF~\cite{zhou2016view} & \xmark & \cmark & \cmark & \xmark \\
    
    CORN~\cite{haeni2020corn}  & \xmark & \cmark & \cmark & \cmark\\
    
    SHARF~\cite{rematasICML21} & \cmark & \cmark & \cmark & \cmark \\
    
    CRGAN~\cite{tian2018cr} & \xmark & \cmark & \cmark & \xmark \\
    
    TBN~\cite{olszewski2019transformable} & \xmark & \cmark & \cmark & \xmark \\
    
    SRN~\cite{sitzmann2019scene} & \xmark & \cmark & \cmark & \cmark \\
    
    PixelNerf~\cite{yu2021cvpr} & \xmark & \cmark & \cmark & \cmark \\
    
    ENR~\cite{dupont2020equivariant} & \xmark & \cmark & \cmark & \xmark \\
    
    VIGAN~\cite{xu2019view} & \xmark & \cmark & \cmark & \cmark \\

    AUTO3D~\cite{liu2020auto3d} & \xmark & \xmark & \cmark & \xmark \\
    \midrule
    Ours & \xmark & \xmark & \xmark & \xmark \\ 
    \bottomrule
  \end{tabular}}
  \caption{Requirements of methods for Novel View Synthesis from a single image. 
  }
  \label{tab:NVS_benchmark}
\end{table}

\section{Related Works}
\subsection{Novel View Synthesis}
Traditional approaches for the NVS problem are based on multi-view geometry ~\cite{debevec1996modeling,fitzgibbon2005image,seitz2006comparison}.
They estimate an explicit 3D representation (e.g., mesh) of the object, then render novel views from it. 
More recently several approaches based on CNNs have been proposed. Some of them, inspired by traditional 3D computer vision, train networks to explicitly estimate a 3D representation of the object such as mesh ~\cite{pontes2018image2mesh}, point-cloud ~\cite{lin2018learning}), voxel map ~\cite{henderson2019learning}, depth map ~\cite{flynn2016deepstereo, xie2016deep3d} and then use it for rendering.
All these methods need some kind of 3D supervision, except for ~\cite{wu2020unsupervised} which however assumes quasi symmetric objects to generate a mesh from a single view.
A different line of work is represented by learning-based approaches, which estimate novel views by directly mapping a source to a target view ~\cite{ji2017deep,dosovitskiy2015learning,zhou2016view, park2017transformation,tian2018cr,tran2017disentangled,sun2018multi, tatarchenko2016multi,xu2019view, yang2015weakly} without explicit estimation of a 3D model. Typically these methods achieve better NVS performance thanks to their ability to hallucinate views even when no geometric information is available (e.g., parts outside of the object mesh).
Among them, we find approaches suitable for a single object instance  ~\cite{lombardi2019neural,sitzmann2019deepvoxels,mildenhall2020nerf} or able to generalize to novel scenes/objects ~\cite{Rhodin_2018_ECCV,sitzmann2019scene,chen2019monocular,worrall2017interpretable,liu2020auto3d, xu2019view, haeni2020corn,yin2020deformvae,tatarchenko2016multi,zhou2016view,olszewski2019transformable,zhu2014multi,Flynn_2019_CVPR}. 
Training an instance-specific model produces higher quality results at the cost of longer training times for each object. General approaches are trained once per category and thus can be more easily deployed on real problems.
However, all the proposed solutions require poses, 3D models or multiple views as training supervision. We report in \cref{tab:NVS_benchmark} a list of relevant learning based methods for novel view synthesis from a single image and their requirements at training or test time.
To the best of our knowledge ours is the first work to relax these assumptions and learn a category-specific network for NVS from natural images only.

\begin{figure*}
    \centering
    \includegraphics[width=0.98\textwidth]{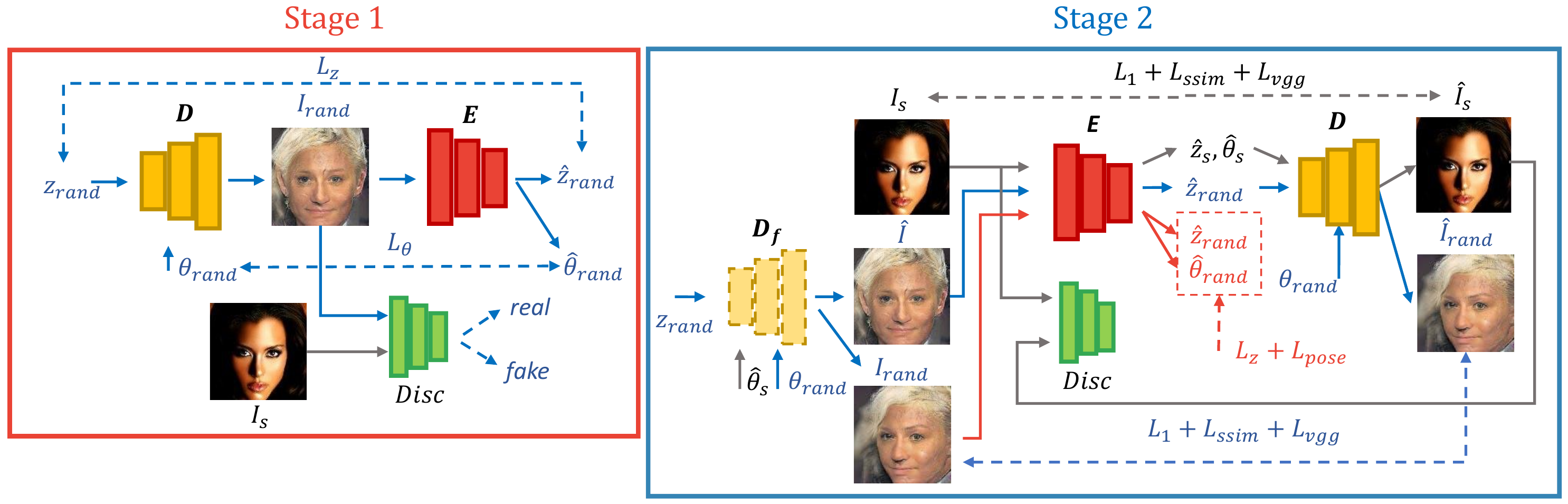}
    \caption{Overview of \algoname{}. During the first stage (left) we train a 3D-aware decoder ($D$) network using a generative formulation while simultaneously training an encoder ($E$) to invert the generation process and regress the latent code $z$ and the pose $\theta$. Then, during the second stage (right), we swap the encoder with the decoder and train the model in an autoencoder fashion to reconstruct real images and self-distill the pre-trained generative model to keep multi-view consistency.}
    \label{fig:framework}
\end{figure*}

\subsection{Inverting Generative Models}
Generative Adversarial Networks (GANs) ~\cite{goodfellow2014generative} have been applied to many computer vision applications, making them the most popular generative models.
Thanks to recent discoveries in terms of architecture ~\cite{brock2018large, Karras_2019_CVPR, Isola_2017_CVPR, Park_2019_CVPR}, loss functions ~\cite{Mao_2017_ICCV, arjovskyICML2017}, and regularization ~\cite{miyato2018spectral, gulrajaniNIPS2017}, GANs are able to synthesize images of impressive realism ~\cite{Karras_2019_CVPR, brock2018large}.
A recent line of work explored the possibility of inverting a pre-existing generator to use it for image manipulation tasks.
These approaches propose strategies to map an image into the generator latent space, and to navigate it to produce variants of the original picture.
Two main families of solutions have been proposed for this task: passing the input image through an encoder neural network ~\cite{kingmaICLR2014, zhuECCV2016} or optimizing a random initial latent code by backpropagation to match the input image ~\cite{Abdal_2019_ICCV, Abdal_2020_CVPR, creswellTNNL2019, zhuECCV2016}. 
Both techniques look for the latent variable that better reconstructs the input image while keeping the generator frozen.
Recent works ~\cite{Abdal_2019_ICCV, Abdal_2020_CVPR, guan2020collaborative, wu2020stylespace, xu2021generative} achieve incredible inversion results by focusing on StyleGAN, which however limits their applicability when used on different architectures.
More general solutions still have limitations on the reconstruction quality.
Indeed, \emph{inverting} a generative model is notoriously difficult due to the complexity and non convexity of the generator latent space ~\cite{zhuECCV2016} or due to the limited variability of objects that can be generated by it.
In our work, along the lines of this research avenue, we propose a method to embed real images of a 3D-aware generative model allowing for identity-preserving transformations in pose (also known as Novel View Synthesis).

Existing GAN inversion works keep the generator frozen in order to maintain the properties of the latent space, making it more difficult to further improve the image reconstruction. 
However, 3D-aware GANs disentangle the pose from the latent space, enabling finetuning of the generator without losing control on the pose.
We propose to use this property during the second stage of our method to finetune the generator and greatly improve the reconstruction quality.

Similar directions have been explored in ConfigNet ~\cite{kowalskiECCV2020} which proposes to use an encoder to embed face images into a disentangled latent space for several attributes  (e.g. color, haircut, pose\dots) allowing a great variety of explicit image manipulations. However, the method needs a 3D synthetic parametric model to generate training data and therefore it has been tested only on faces.
Our method instead does not need any form of parametric model, but can be trained on a simple collection of natural images.

\subsection{3D-Aware Neural Image Synthesis}
Generative models have recently been extended with different forms of 3D representations embedded directly in the network architecture ~\cite{Liao_2020_CVPR, alhaijaACCV2018, Nguyen-Phuoc_2019_ICCV, wangECCV2016, zhuNIPS2018, schwarz2020graf} to synthesize objects from multiple views. 
The majority of methods require 3D supervision ~\cite{wangECCV2016, zhuNIPS2018} or 3D information as input ~\cite{Oechsle_2019_ICCV,alhaijaACCV2018}. 
~\cite{henzler2019platonicgan, Nguyen-Phuoc_2019_ICCV, schwarz2020graf} train generative models without supervision from only a collections of natural images.
~\cite{henzler2019platonicgan} learns a textured 3D voxel representation from 2D images using a differentiable render, while HoloGAN ~\cite{Nguyen-Phuoc_2019_ICCV} and some related works ~\cite{Liao_2020_CVPR, nguyenphuoc2020blockgan, Niemeyer_2020_CVPR} learn a low-dimensional 3D feature combined with a 3D-to-2D projection. GRAF ~\cite{schwarz2020graf} leverages neural radiance fields ~\cite{mildenhall2020nerf} conditioned on a random variable to enable generation of novel objects.
We build on these recent findings and extend 3D aware generative models to perform novel view synthesis from natural images.

\section{Method}
The key concept behind the proposed framework is to exploit the implicit knowledge of 3D-aware generative models in order to perform novel view synthesis. 
\cref{fig:framework} shows an overview of the proposed approach.

Defined as $\theta_s$ and $\theta_t$ a source and a target unknown view-points, our goal is to synthesize a novel view $I_t$ from $\theta_t$ given only a single image $I_s$ representing an object from the view $\theta_s$.
Our framework, namely \algoname{}, is composed of two main components: an encoder $E$ and a decoder $D$.
$E$ takes as input $I_s$ and estimates a global latent representation $z_s$ jointly with the source view $\theta_s$.
$D$, instead, is a a 3D-aware generative model: given a global latent object representation $z$, and a target view $\theta$, it can generate an image depicting the object in the requested view.
However, differently from the generative literature, we aim to use $D$ to generate novel views of a specific \emph{real} object rather than a \emph{generated} one. 

%Thus, given $z_s$ and $\theta_t$, $D$ produces the novel view $I_t$.

\subsection{Stage 1: Generative Training}\label{sec:stage1}
A naive solution for our problem could be to train an autoencoder to reconstruct images, \ie{} $\theta_t = \theta_s$.
A similar solution was used by ~\cite{olszewski2019transformable} but relying on multi-view supervision at training time. However, in our scenario we cannot enforce multi-view consistency due to the lack of annotations, therefore the network is free to learn incomplete representations that are good to reconstruct the input image but cannot be used to generate novel views, \eg it might learn to ignore the pose. We will show experimental results to support this claim in \cref{ss:two_stage_ablation}.
To overcome this problem, we decide to train our framework in a generative manner in a first training step: \textit{Stage 1}.
In particular, we train $D$ to create realistic views starting from randomly sampled latent representations, $z_{rand}$, and poses, $\theta_{rand}$. This stage helps $D$ to learn a good initial representation for the object category. 
Recent 3D-aware generators ~\cite{Nguyen-Phuoc_2019_ICCV, schwarz2020graf} can be trained without using any kind of supervision, therefore for Stage 1 we use only natural images.
We train our generative model using a standard adversarial loss $L_{adv}$ which involves the use of a discriminator, $Disc$.
Jointly with $D$, we also train an encoder $E$ to map a generated image to its latent space and pose.

Overall, we use $D$ to generate an image $I_{rand} = D(z_{rand},\theta_{rand})$ from a randomly sampled latent code and pose; then we optimize $E$ to invert the generation process, estimating the view $\hat{\theta}_{rand}$  and latent vector $\hat{z}_{rand}$ by minimizing the following losses:

\begin{equation}
    L_z = \E{|| z_{rand} - \hat{z}_{rand}||}
    \label{eq:z_cycle}
\end{equation}
\begin{equation}
    L_\theta = \E{||\theta_{rand} - \hat{\theta}_{rand}||}
    \label{eq:view_cycle}
\end{equation}

\subsection{Stage 2: Reconstruction Autoencoder}
\label{sec:stage2}
Given a real image, we could use $E$ to estimate a $z$ and $pose$ that would project it into the learned generator latent space and, in turn, use $D$ for image reconstruction and NVS (changing $pose$ while keeping $z$). 
However, this \textit{GAN inversion} task is notoriously difficult due to the non-convexity of the generator latent space ~\cite{zhuECCV2016}. 
In practice very often the $z$ and $pose$ predicted for real images generate very poor reconstructions (see \cref{sec:gan_inversion_exp}).
We argue that we might obtain much better results by finetuning the weights of the generator differently from standard inversion approaches.
Typically, these methods keep the generator frozen to preserve the structure of the generative space and be able to perform image editing tasks in it.
However, in our case the architecture of $D$ disentangles the $pose$ from $z$ by construction, therefore even if we finetune the weights of the generator we could always retain the ability to modify the pose of the generated images.
Thus, in \textit{Stage 2}, we swap the decoder with the encoder, rebuilding the original autoencoder structure, and we train the whole system to reconstruct real images in a end-to-end fashion.

Summarizing, given a source real image $I_s$ we propagate it through the encoder $E$ to get the corresponding $E(I_s)=(\hat{z}_s, \hat{\theta}_s)$, \ie{,} the predicted latent vector and view respectively. 
At this point, we use $D$ to reconstruct the original image $\hat{I}_s = D(\hat{z}_s$, $\hat{\theta}_s)$ by starting from the pre-trained $D$ and $E$ from Stage 1.
We train our system with a combination of pixel wise $L_2$, perceptual ~\cite{isola2017image} $L_{vgg}$, and SSIM ~\cite{wang2004image} $L_{ssim}$ losses, defined as follows:

\begin{equation}
\label{eq:autoencoder_l1}
    L_2 = \E{||I_s - \hat{I}_s||} \\
\end{equation}
\begin{equation}
\label{eq:autoencoder_vgg}
    L_{vgg} = \sum_{i}{\E{||V_i(I_s) - V_i(\hat{I}_s)||}}
\end{equation}
\begin{equation}
\label{eq:autoencoder_ssim}
    L_{ssim} = \E{(1 - SSIM(I_s, \hat{I}_s))}
\end{equation}

where $V_i$ are features extracted at the $i$-th layer of a VGG16 ~\cite{Simonyan15} backbone pre-trained on ImageNet ~\cite{dengICCV2009}.
We experimentally achieved best reconstruction results by using the output of the \textit{block2\_conv2} layer.
Moreover, to increase the quality and details of the reconstruction, we follow the example of ~\cite{Isola_2017_CVPR} and we train the generator adversarially with $Disc$ using the DCGAN ~\cite{radfordICLR2016} loss, $L_{adv}$, computed using $I_s$ as real samples and $\hat{I}_s$ as fake samples.
Since our generator is  pre-trained, to stabilize the adversarial learning, we start from the pre-trained $Disc$ of Stage 1.

\subsection{Stage 2: Distilling the Generative Knowledge}
\label{ss:distilling}
As supported by experiments in \cref{ss:two_stage_ablation}, we verified that by simply finetuning the network for reconstruction on all images, it will collapse to an inconsistent 3D representation that cannot be used to synthesize novel views by changing $\theta$.
Nevertheless, we know that $D$ has a consistent 3D representation at the end of Stage 1. 
Thus, we force our network to preserve such a crucial property by \textit{self-distilling} the knowledge from a frozen version of itself, namely $D_f$.
To do so, we train the model on a set of images generated by $D_f$.
For these images we can have both ``labels'' for their latent code and pose, as well as multiple-views of the same object.
We use all these information to train the encoder and decoder with direct supervision.
Thus, for the training of $E$, given a random latent vector $z_{rand}$ and a random view 
$\theta_{rand}$ we generate an image from the frozen decoder $D_f$ as $I_{rand} = D_f(z_{rand},\theta_{rand})$.
Then, we can predict $\hat{z}_{rand}$ and $\hat{\theta}_{rand}$ from $E(I_{rand})$ and apply the consistency losses described in \cref{eq:z_cycle} and \cref{eq:view_cycle}.
For the training of $D$, given $z_{rand}$ and $\theta_s$, we use $D_f$ to generate a novel view of the same object, $\hat{I}$.
Instead of applying the standard autoencoder loss, we use $E$ and $D$ to reconstruct a novel view of the object and provide direct supervision on it.
By doing so we preserve the consistency of 3D-representations by maintaining good geometric priors.
To do so, the encoder predicts the latent vector $\hat{z}_{rand}$ from $\hat{I}$ and the decoder generates a novel view as $\hat{I}_{rand} = D(\hat{z}_{rand}, \theta_{rand})$.
Then, we apply the previous reconstruction losses to the novel view:

\begin{equation}
    L_2^{gen} = \E{||\hat{I}_{rand} - I_{rand}||} \\
\end{equation}
\begin{equation}
    L_{vgg}^{gen} = \sum_{i}{\E{||V_i(\hat{I}_{rand}) - V_i( I_{rand})||}}
\end{equation}
\begin{equation}
    L_{ssim}^{gen} = \E{(1 - SSIM(\hat{I}_{rand}, I_{rand}))}
\end{equation}

At test time given an input image $I$ of an object we can use $E$ to predict the corresponding $(z,\theta)$, by keeping $z$ fixed and changing $\theta$ we can easily synthesize novel views of the specific object depicted in $I$.

\subsection{Self-Adaptation by finetuning}\label{sec:finetuning}
Finally, we propose an optional final step where we employ one-shot learning to finetune our network on a specific target image with a self-supervised protocol.
We use the set of losses $L_2$, $L_{ssim}$, $L_{vgg}$, and $L_{adv}$ to optimize the parameters of $D$ and the value of $(z, \theta)$ by back-propagation starting from the initial values provided by $E$.
This step is optional but we will show in \cref{sec:gan_inversion_exp} how it can improve quality by adding some fine details to the reconstructed images without downgrading the ability to generate novel views.

\begin{table*}[ht]
    \centering
    \scalebox{0.9}{
    \begin{tabular}{cccccccc}
         Name & Type & \# Training/Val/Test & Resolution & Azimuth & Elevation & Scaling \\
         \toprule
         CelebA~\cite{liu2015faceattributes} &  Real & 162770 / 19867 / 19962 & 128x128 & 220$^{\circ}$-320$^{\circ}$ & 70$^{\circ}$-110$^{\circ}$ & 1.0\\
         Real Cars ~\cite{yang2015large} & Real & 95410 / 13633 / 27267 & 128x128 & 0$^{\circ}$-360$^{\circ}$ & 60$^{\circ}$-95$^{\circ}$ & 1.0-1.5 \\
         Shapenet-Cars ~\cite{shapenet2015} & Synthetic & 125928 / 18000 / 35976 & 128x128 & 0$^{\circ}$-360$^{\circ}$ & 25$^{\circ}$-30$^{\circ}$ & 1.0-1.5  \\
         Shapenet-Sofa ~\cite{shapenet2015} & Synthetic & 53304 / 7608 / 15239 & 128x128 & 0$^{\circ}$-360$^{\circ}$ & 25$^{\circ}$-30$^{\circ}$ & 1.0-1.5 \\
         \bottomrule
    \end{tabular}}
    \caption{Additional Datasets Information}
    \label{tab:datasets}
\end{table*}
\section{Experimental Settings}
\label{sec:details}

\subsection{Network Architectures.}
We focus on experiments using HoloGAN ~\cite{Nguyen-Phuoc_2019_ICCV} as $D$, because of its lower computational requirements w.r.t. neural radiance field generators ~\cite{schwarz2020graf}. 
The lightweight architecture of HoloGAN allows us to employ image-wide perceptual losses which are extremely important to achieve better performance (see \cref{sec:adversarial}). 
For neural radiance fields models the generation of a full frame for each training sample
would prohibitively increase the computational costs. 
However, we do recognize the interesting properties of NeRF-based methods and for this reason we report preliminary results using GRAF as generator in \cref{sec:graf}.
We modify the basic architecture of HoloGAN by using for up-sampling 3x3 convolutions followed by a bilinear sampling instead of deconvolutions to decrease the well known grid artifacts problem ~\cite{odena2016deconvolution}. 
HoloGAN encodes the pose $\theta$ as the azimuth and elevation of the object with respect to a canonical reference system which is learnt directly from the network, assuming the object being at the center of the scene. Moreover, it uses a scale parameter representing the object dimension which is inversely related to the distance between the object and the virtual camera.
We use the same parametrization and train $E$ to estimate $\theta$ as azimuth, elevation, and scale. 
$E$ is composed of three stride-2 residual blocks with reflection padding followed by two heads for pose and $z$ regression. 
The pose head is composed of a sequence of 3x3 convolution, average pooling, fully connected and sigmoid layers. The estimated pose is remapped from [0,1] into a given range depending on the datasets. 
The $z$ head has a similar architecture except for the tanh activation at the end. $z$ is a 1D array of shape $128$ except for Real Cars when we use shape $200$. During Stage 1 we sample $z$ from a $[0,1]$ uniform distribution.

\subsection{Training Details.}
Our pipeline is implemented using Tensorflow ~\cite{tensorflow2015-whitepaper} and trained on a NVIDIA 1080 Ti with 11 GB of memory. During Stage 1 we use a procedure similar to the one described in ~\cite{Nguyen-Phuoc_2019_ICCV}.
We train our networks for 50 and 30 epochs for Stage 1 and 2 respectively. We use Adam ~\cite{adam} with $\beta_1$ 0.5 and $beta_2$ 0.999. Initial learning rate is $5*10^{-5}$ and after 25 epochs we start to linearly decay it. During the optional third stage we finetune for 100 steps with learning rate $10^{-4}$.

\subsection{Datasets.}
We evaluate our framework on ShapeNet ~\cite{shapenet2015}, CelebA ~\cite{liu2015faceattributes} and Real Cars ~\cite{yang2015large}. 
We train at resolution 128x128 for all datasets.
Even when available, we did not make use of any annotations during training. 
ShapeNet ~\cite{shapenet2015} is a synthetic dataset of 3D models belonging to various categories. We tested our method in the \textit{cars} and \textit{sofa} categories using renderings from ~\cite{choy20163d} as done in ~\cite{xu2019view, liu2020auto3d}. For each category, following the standard splits defined in  ~\cite{xu2019view,liu2020auto3d}, we use 70\% of the models for training, 20\% for testing and 10\% to validate our approach. Even though the dataset provides multiple views of each object, we did not use this information at training time since our method does not require any multi-view assumption.
CelebA ~\cite{liu2015faceattributes} is a dataset composed of 202599 images of celebrity faces. 
We use the aligned version of the dataset, center cropping the images.
Real Cars ~\cite{yang2015large} is a large scale dataset of real cars composed of 136726 images.
For real datasets we split them in 80\% for training and 20\% for testing.
We make use of the bounding box provided from the dataset to crop out a square with the same center and size equal to the largest side of the car bounding box. If the square exceeds image dimensions, we crop the largest possible square. Then, we resize images at 128x128.

Finally, in \cref{tab:datasets} we report additional information on our training datasets.

\begin{table}
\centering
\scalebox{0.7}{
\centering
  \begin{tabular}{c|ccc|c|cc}
    \toprule
     & \multicolumn{3}{c|}{Train} & Test & \multicolumn{2}{c}{Sofa} \\
     \midrule
    Method & 3D & Pose & Multiview & Source Pose & $L_1\downarrow$ & SSIM$\uparrow$ \\
    \midrule
    \cite{tatarchenko2016multi} & \cmark & \cmark & \cmark & \xmark & 17.52 & 0.73\\
    \cite{zhou2016view} & \xmark & \cmark & \cmark & \xmark & 13.26 & 0.77 \\
    \cite{xu2019view}& \xmark & \cmark & \cmark & \cmark & \textbf{10.13} & \textbf{0.83} \\
    \cite{liu2020auto3d} & \xmark & \xmark & \cmark & \xmark & 10.30 & 0.82 \\
    \midrule
    Ours & \xmark & \xmark & \xmark & \xmark & 13.83 & 0.72\\
    \bottomrule
  \end{tabular}}
  \caption{Evaluation of NVS using a single input image on the ShapeNet-Sofa dataset. The best results are highlighted in bold. When computing the $L_1$ error, pixel values are in range of [0, 255]. 
  }
  \label{tab:NVS}
\end{table}

\begin{figure}
	\setlength{\tabcolsep}{0.5pt}
	\renewcommand{\arraystretch}{0.5} 
	\centering
	\setlength{\fboxrule}{2pt} 
	\setlength{\fboxsep}{0pt} 
	\scalebox{1}{
	\begin{tabular}{cc|ccccccc}
	 & {\scriptsize Input.} & \multicolumn{7}{|c}{\scriptsize $\theta$ Interpolation} \\
	 \rotatebox{90}{\hspace{0.5em} \scriptsize Ours} &
	 \includegraphics[width=0.115\linewidth]{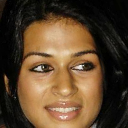} & \includegraphics[width=0.115\linewidth]{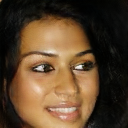} & \includegraphics[width=0.115\linewidth]{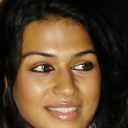} & \includegraphics[width=0.115\linewidth]{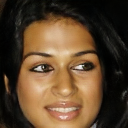} & \includegraphics[width=0.115\linewidth]{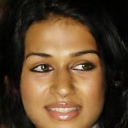} & \includegraphics[width=0.115\linewidth]{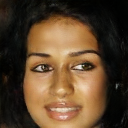} & \includegraphics[width=0.115\linewidth]{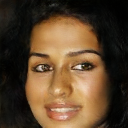} & \includegraphics[width=0.115\linewidth]{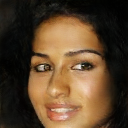} 
	 \\
	 \rotatebox{90}{\scriptsize CR-GAN} &
     \includegraphics[width=0.115\linewidth]{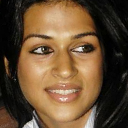} & \includegraphics[width=0.115\linewidth]{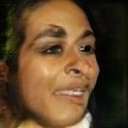} & \includegraphics[width=0.115\linewidth]{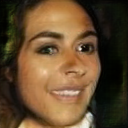} & \includegraphics[width=0.115\linewidth]{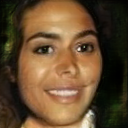} & \includegraphics[width=0.115\linewidth]{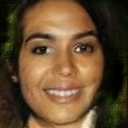} & \includegraphics[width=0.115\linewidth]{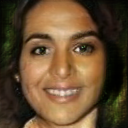} & \includegraphics[width=0.115\linewidth]{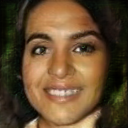} & \includegraphics[width=0.115\linewidth]{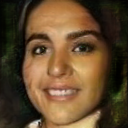} \\
     \midrule
 	 \rotatebox{90}{\hspace{0.4em} \scriptsize Ours} &
     \includegraphics[width=0.115\linewidth]{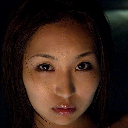} & \includegraphics[width=0.115\linewidth]{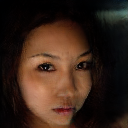} & \includegraphics[width=0.115\linewidth]{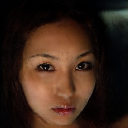} & \includegraphics[width=0.115\linewidth]{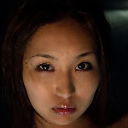} & \includegraphics[width=0.115\linewidth]{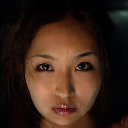} & \includegraphics[width=0.115\linewidth]{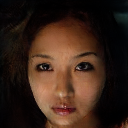} & \includegraphics[width=0.115\linewidth]{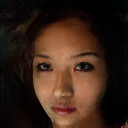} & \includegraphics[width=0.115\linewidth]{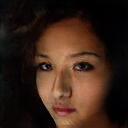} \\
	 \rotatebox{90}{\scriptsize CR-GAN} &
     \includegraphics[width=0.115\linewidth]{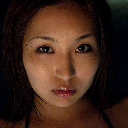} & \includegraphics[width=0.115\linewidth]{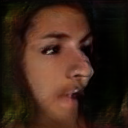} & \includegraphics[width=0.115\linewidth]{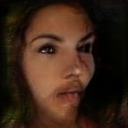} & \includegraphics[width=0.115\linewidth]{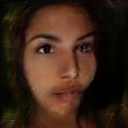} & \includegraphics[width=0.115\linewidth]{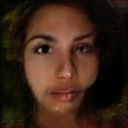} & \includegraphics[width=0.115\linewidth]{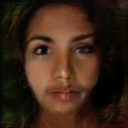} & \includegraphics[width=0.115\linewidth]{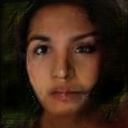} & \includegraphics[width=0.115\linewidth]{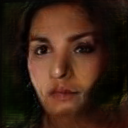} \\
	\end{tabular}}
	\caption{Qualitative comparison vs CR-GAN \cite{tian2018cr} on CelebA. For each input, top row is our method, bottom row is CR-GAN.}
	\label{fig:comparison_crgan}
\end{figure}

\section{Experimental Results}

\subsection{Novel View Synthesis}
\label{ss:nvs}

\textbf{Quantitative Comparison.}
We investigate the performance of our Stage 2 (no image-specific finetuning) framework for NVS from object rotation.
We compare mainly against Auto3D ~\cite{liu2020auto3d} because it is the work with the requirements closer to ours (even if requires multi-view images at training time while we don't).
We follow the protocol defined by Auto3D and we show the performance of single image NVS for the class Sofa in ShapeNet ~\cite{shapenet2015}. We do not report results for the classes Bench and Chairs because our generative model based on HoloGAN during stage 1 failed to find a meaningful 3D representation for those classes (more on this in \cref{sec:limitations}).
For each pair of views of an object, $(I_s, I_t)$, we evaluate the L1 (lower is better) and SSIM ~\cite{wang2004image} (higher is better) reconstruction metrics and we report results in \cref{tab:NVS}.
Given an input image $I_s$, we first estimate the global latent vector $z_s = E(I_s)$. 
Then, we generate the target view $I_t = D(z_s, \theta_t)$ using the ground-truth target pose $\theta_t$. %
When computing the $L_1$ error we use pixel values in range $[0,255]$.
Since the pose estimated by our network is aligned to a learnt reference system which may differ from the one of ShapeNet, we train a linear regressor on the training set to map the ground-truth reference system into the learnt one. We use this mapping at test time to compare our method with other works. However, this is not needed in practise, and it could be a source of error during the evaluation.
Our method achieves results comparable to MV3D ~\cite{tatarchenko2016multi} with a lower L1 error but with a slightly lower SSIM. 
Compared to the other works we achieve slightly worse performance, however, as highlighted in \cref{tab:NVS_benchmark}, all competitors require multiple views of the same object or ground-truth poses ~\cite{tatarchenko2016multi, zhou2016view, xu2019view}, while we do not.

\textbf{Qualitative Results.}
In \cref{fig:comparison_crgan} we extend the test to a real dataset and qualitatively compare against CR-GAN ~\cite{tian2018cr} on the test set of CelebA. 
For CR-GAN we use the original code implementation and pre-trained weights available online and we pre-process images as described in the paper, which is different from our pre-processing, leading to slightly different inputs.
Our method consistently achieves better results since it better preserves the identity of the input image and the 3D structure, especially for the extreme views from the side. 
We wish to highlight that CR-GAN was pre-trained with multi-view and pose supervision on the 300wLP dataset ~\cite{zhu2016face} before being fine tuned on CelebA.
Moreover CR-GAN addresses pose estimation as a classification task, therefore it can generate only a fixed number of poses, while our method allows to sample continuous poses.

 \begin{figure}
	\setlength{\tabcolsep}{0.5pt}
	\renewcommand{\arraystretch}{0.5} 
	\centering
	\setlength{\fboxrule}{2pt} 
	\setlength{\fboxsep}{0pt} 
	\scalebox{1}{
	\begin{tabular}{cc|ccccccc}
	& {\scriptsize Input.} & \multicolumn{7}{|c}{\scriptsize $\theta$ Interpolation} \\
    \rotatebox{90}{\hspace{0.5em}\scriptsize Ours} &
	\includegraphics[width=0.115\linewidth]{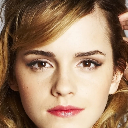} & \includegraphics[width=0.115\linewidth]{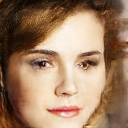} & \includegraphics[width=0.115\linewidth]{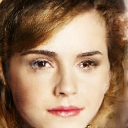} & \includegraphics[width=0.115\linewidth]{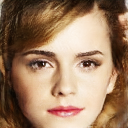} & \includegraphics[width=0.115\linewidth]{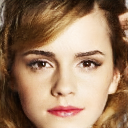} & \includegraphics[width=0.115\linewidth]{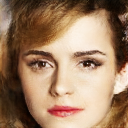} & \includegraphics[width=0.115\linewidth]{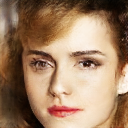} & \includegraphics[width=0.115\linewidth]{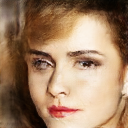} \\ 
     \rotatebox{90}{\hspace{0.2em}\scriptsize ConfigNet} &
    \includegraphics[width=0.115\linewidth]{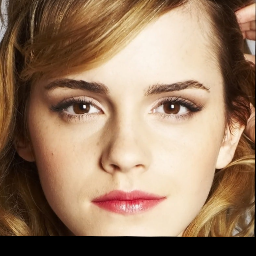} & \includegraphics[width=0.115\linewidth]{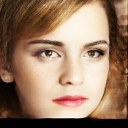} & \includegraphics[width=0.115\linewidth]{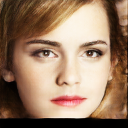} & \includegraphics[width=0.115\linewidth]{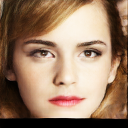} & \includegraphics[width=0.115\linewidth]{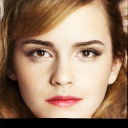} & \includegraphics[width=0.115\linewidth]{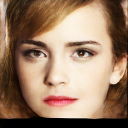} & \includegraphics[width=0.115\linewidth]{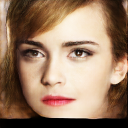} & \includegraphics[width=0.115\linewidth]{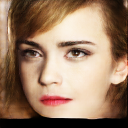} \\
    \rotatebox{90}{\hspace{0.5em}\scriptsize Ours} &
    \includegraphics[width=0.115\linewidth]{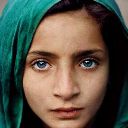} & \includegraphics[width=0.115\linewidth]{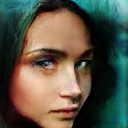} & \includegraphics[width=0.115\linewidth]{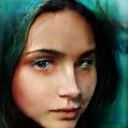} & \includegraphics[width=0.115\linewidth]{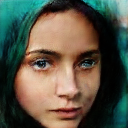} & \includegraphics[width=0.115\linewidth]{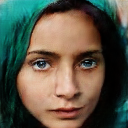} & \includegraphics[width=0.115\linewidth]{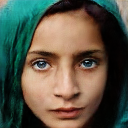} & \includegraphics[width=0.115\linewidth]{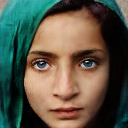} & \includegraphics[width=0.115\linewidth]{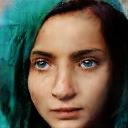} \\
    \rotatebox{90}{\hspace{0.2em} \scriptsize ConfigNet} &
    \includegraphics[width=0.115\linewidth]{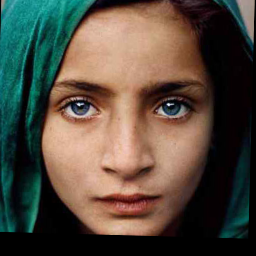} & \includegraphics[width=0.115\linewidth]{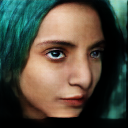} & \includegraphics[width=0.115\linewidth]{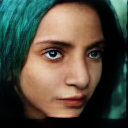} & \includegraphics[width=0.115\linewidth]{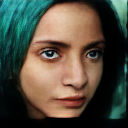} & \includegraphics[width=0.115\linewidth]{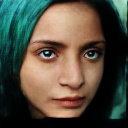} & \includegraphics[width=0.115\linewidth]{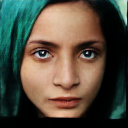} & \includegraphics[width=0.115\linewidth]{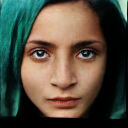} & \includegraphics[width=0.115\linewidth]{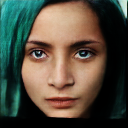} \\
	\end{tabular}}
	\caption{Qualitative comparison against ConfigNet \cite{kowalskiECCV2020} on downloaded web images. For each input image, the top row is our method, while the bottom row is ConfigNet.}
	\label{fig:comparison_cfgnet}
\end{figure}

In \cref{fig:comparison_cfgnet}, we compare our results against ConfigNet ~\cite{kowalskiECCV2020} on random portrait images downloaded from the web. 
For ConfigNet we align the faces using the pre-processing based on OpenFace~\cite{amos2016openface} as in the original paper, thus the inputs are slightly different (e.g. thin black borders for ConfigNet). 
We used the original code implementation available online and the pre-trained weights on FFHQ ~\cite{Karras_2019_CVPR}. 
We notice that our network achieves comparable results, slightly better for poses closer to the input image, while worse for far poses. 
However, differently from our method, ConfigNet requires large synthetic datasets with labels for pose and several semantic attributes for an initial supervised training, while we do not have this type of constraint. 
We argue that while it is remarkable that high quality rendering of faces can be used to pre-train a model with supervision, it is hard to think of extending the same procedure to any other object categories we might be interested in, since it requires curated 3D modeling and expensive rendering times.
Our model provides comparable performance without these requirements since we only need a collection of natural images.  

 \begin{figure*}[ht]
	\setlength{\tabcolsep}{0.5pt}
	\renewcommand{\arraystretch}{0.5} 
	\centering
	\setlength{\fboxrule}{2pt} 
	\setlength{\fboxsep}{0pt} 
	\scalebox{0.98}{
	\begin{tabular}{c|cccccc||c|cccccc}
	\multicolumn{7}{c||}{\scriptsize CelebA} & \multicolumn{7}{c}{\scriptsize Real Cars}  \\
	\midrule
	{\scriptsize Input.} & \multicolumn{6}{c||}{\scriptsize $\theta$ Interpolation} & {\scriptsize Input.} & \multicolumn{6}{c}{\scriptsize $\theta$ Interpolation} \\
	
         \includegraphics[width=0.07\textwidth]{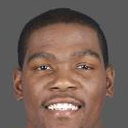} & \includegraphics[width=0.07\textwidth]{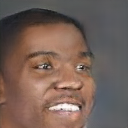} & \includegraphics[width=0.07\textwidth]{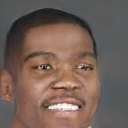} & \includegraphics[width=0.07\textwidth]{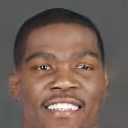} & \includegraphics[width=0.07\textwidth]{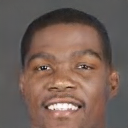} & \includegraphics[width=0.07\textwidth]{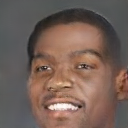} & \includegraphics[width=0.07\textwidth]{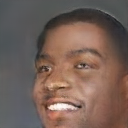} &
         \includegraphics[width=0.07\textwidth]{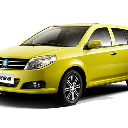} & \includegraphics[width=0.07\textwidth]{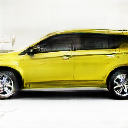} & \includegraphics[width=0.07\textwidth]{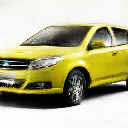} & \includegraphics[width=0.07\textwidth]{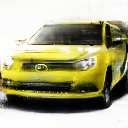} & \includegraphics[width=0.07\textwidth]{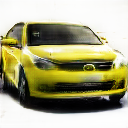} & \includegraphics[width=0.07\textwidth]{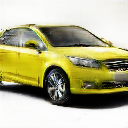} & \includegraphics[width=0.07\textwidth]{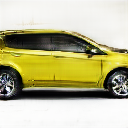} \\
         
         \includegraphics[width=0.07\textwidth]{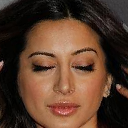} & \includegraphics[width=0.07\textwidth]{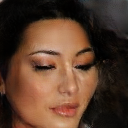} & \includegraphics[width=0.07\textwidth]{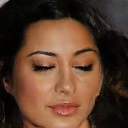} & \includegraphics[width=0.07\textwidth]{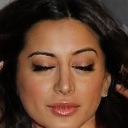} & \includegraphics[width=0.07\textwidth]{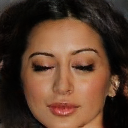} & \includegraphics[width=0.07\textwidth]{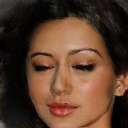} & \includegraphics[width=0.07\textwidth]{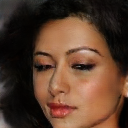} &
         \includegraphics[width=0.07\textwidth]{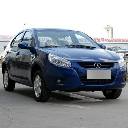} & \includegraphics[width=0.07\textwidth]{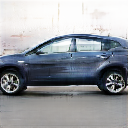} & \includegraphics[width=0.07\textwidth]{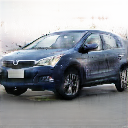} & \includegraphics[width=0.07\textwidth]{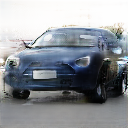} & \includegraphics[width=0.07\textwidth]{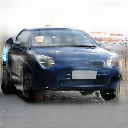} & \includegraphics[width=0.07\textwidth]{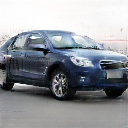} & \includegraphics[width=0.07\textwidth]{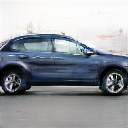} \\
         
         \includegraphics[width=0.07\textwidth]{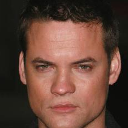} & \includegraphics[width=0.07\textwidth]{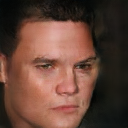} & \includegraphics[width=0.07\textwidth]{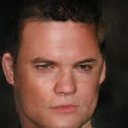} & \includegraphics[width=0.07\textwidth]{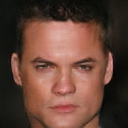} & \includegraphics[width=0.07\textwidth]{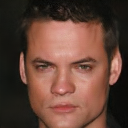} & \includegraphics[width=0.07\textwidth]{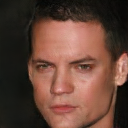} & \includegraphics[width=0.07\textwidth]{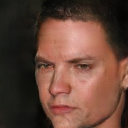} &
         \includegraphics[width=0.07\textwidth]{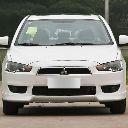} & \includegraphics[width=0.07\textwidth]{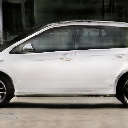} & \includegraphics[width=0.07\textwidth]{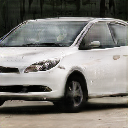} & \includegraphics[width=0.07\textwidth]{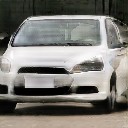} & \includegraphics[width=0.07\textwidth]{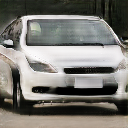} & \includegraphics[width=0.07\textwidth]{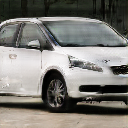} & \includegraphics[width=0.07\textwidth]{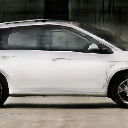} \\
         
         \includegraphics[width=0.07\textwidth]{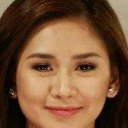} & \includegraphics[width=0.07\textwidth]{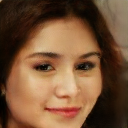} & \includegraphics[width=0.07\textwidth]{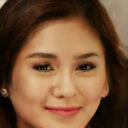} & \includegraphics[width=0.07\textwidth]{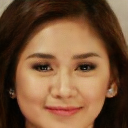} & \includegraphics[width=0.07\textwidth]{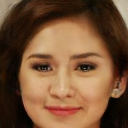} & \includegraphics[width=0.07\textwidth]{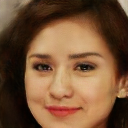} & \includegraphics[width=0.07\textwidth]{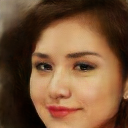} &
         \includegraphics[width=0.07\textwidth]{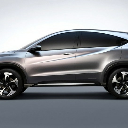} & \includegraphics[width=0.07\textwidth]{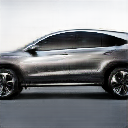} & \includegraphics[width=0.07\textwidth]{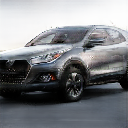} & \includegraphics[width=0.07\textwidth]{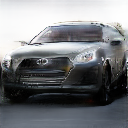} & \includegraphics[width=0.07\textwidth]{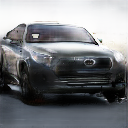} & \includegraphics[width=0.07\textwidth]{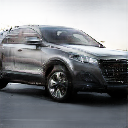} & \includegraphics[width=0.07\textwidth]{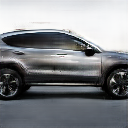} \\
         \midrule
         \multicolumn{7}{c||}{\scriptsize ShapeNet Cars} & \multicolumn{7}{c}{\scriptsize ShapeNet Sofa}  \\
    	 \midrule
    	 {\scriptsize Input.} & \multicolumn{6}{c||}{\scriptsize $\theta$ Interpolation} & {\scriptsize Input.} & \multicolumn{6}{c}{\scriptsize $\theta$ Interpolation} \\
         \includegraphics[width=0.07\textwidth]{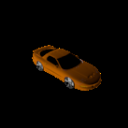} & \includegraphics[width=0.07\textwidth]{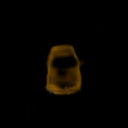} & \includegraphics[width=0.07\textwidth]{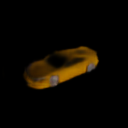} & \includegraphics[width=0.07\textwidth]{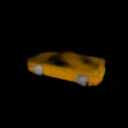} & \includegraphics[width=0.07\textwidth]{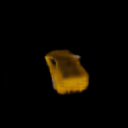} & \includegraphics[width=0.07\textwidth]{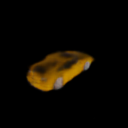} & \includegraphics[width=0.07\textwidth]{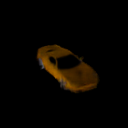} &
         \includegraphics[width=0.07\textwidth]{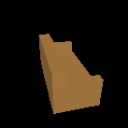} & \includegraphics[width=0.07\textwidth]{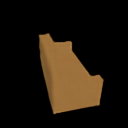} & \includegraphics[width=0.07\textwidth]{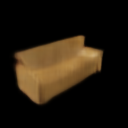} & \includegraphics[width=0.07\textwidth]{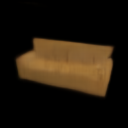} & \includegraphics[width=0.07\textwidth]{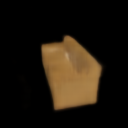} & \includegraphics[width=0.07\textwidth]{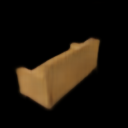} & \includegraphics[width=0.07\textwidth]{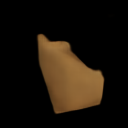} \\

         \includegraphics[width=0.07\textwidth]{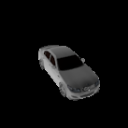} & \includegraphics[width=0.07\textwidth]{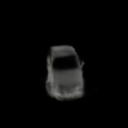} & \includegraphics[width=0.07\textwidth]{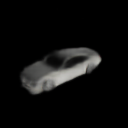} & \includegraphics[width=0.07\textwidth]{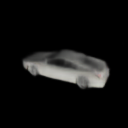} & \includegraphics[width=0.07\textwidth]{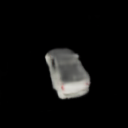} & \includegraphics[width=0.07\textwidth]{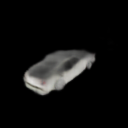} & \includegraphics[width=0.07\textwidth]{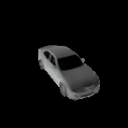} &
         \includegraphics[width=0.07\textwidth]{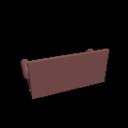} & \includegraphics[width=0.07\textwidth]{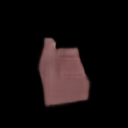} & \includegraphics[width=0.07\textwidth]{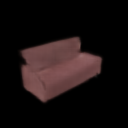} & \includegraphics[width=0.07\textwidth]{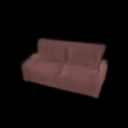} & \includegraphics[width=0.07\textwidth]{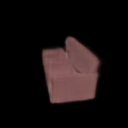} & \includegraphics[width=0.07\textwidth]{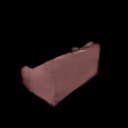} & \includegraphics[width=0.07\textwidth]{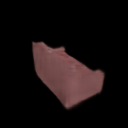} \\
         
         \includegraphics[width=0.07\textwidth]{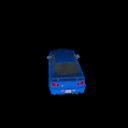} & \includegraphics[width=0.07\textwidth]{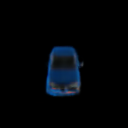} & \includegraphics[width=0.07\textwidth]{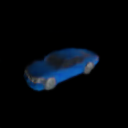} & \includegraphics[width=0.07\textwidth]{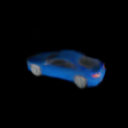} & \includegraphics[width=0.07\textwidth]{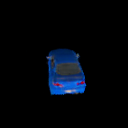} & \includegraphics[width=0.07\textwidth]{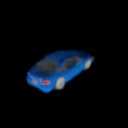} & \includegraphics[width=0.07\textwidth]{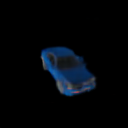} &
         \includegraphics[width=0.07\textwidth]{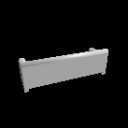} & \includegraphics[width=0.07\textwidth]{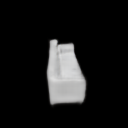} & \includegraphics[width=0.07\textwidth]{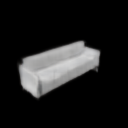} & \includegraphics[width=0.07\textwidth]{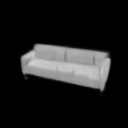} & \includegraphics[width=0.07\textwidth]{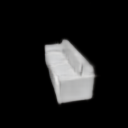} & \includegraphics[width=0.07\textwidth]{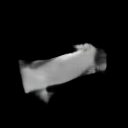} & \includegraphics[width=0.07\textwidth]{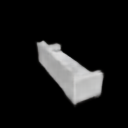} \\
           
         \includegraphics[width=0.07\textwidth]{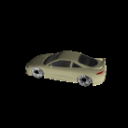} & \includegraphics[width=0.07\textwidth]{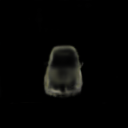} & \includegraphics[width=0.07\textwidth]{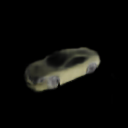} & \includegraphics[width=0.07\textwidth]{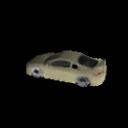} & \includegraphics[width=0.07\textwidth]{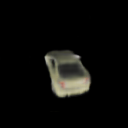} & \includegraphics[width=0.07\textwidth]{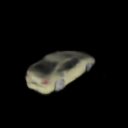} & \includegraphics[width=0.07\textwidth]{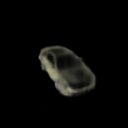} &
         \includegraphics[width=0.07\textwidth]{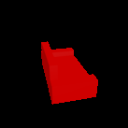} & \includegraphics[width=0.07\textwidth]{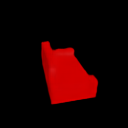} & \includegraphics[width=0.07\textwidth]{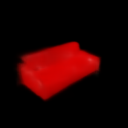} & \includegraphics[width=0.07\textwidth]{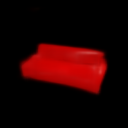} & \includegraphics[width=0.07\textwidth]{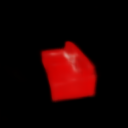} & \includegraphics[width=0.07\textwidth]{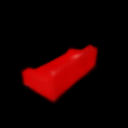} & \includegraphics[width=0.07\textwidth]{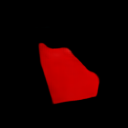} \\
	\end{tabular}}
	\caption{Qualitative results on CelebA \cite{liu2015faceattributes}, Real Cars \cite{yang2015large}, ShapeNet-Cars \cite{shapenet2015}, ShapeNet-Sofa \cite{shapenet2015}.}
	\label{fig:qualitatives}
\end{figure*}

Finally, in \cref{fig:qualitatives} we show additional qualitative results of our method after finetuning for CelebA ~\cite{liu2015faceattributes}, Real Cars ~\cite{yang2015large}, Shapenet-Cars ~\cite{shapenet2015} and Shapenet-Sofa ~\cite{shapenet2015}.

\begin{table}[t]
\centering
\centering
\scalebox{0.98}{
  \begin{tabular}{l|ccc}
  \toprule
  Method & L1 & SSIM & PSNR \\
  \midrule
  ~\cite{creswellTNNL2019} & 65.79 & 0.25 & 10.07 \\
  ~\cite{zhuECCV2016} & 31.18 & 0.40 & 15.43 \\
  Stage 2 & 23.83 & 0.50 & 18.01 \\
  Stage 2 + Finetune & 5.23 & 0.90 & 31.00 \\
  \bottomrule
  \end{tabular}}
  \caption{Reconstruction quality on CelebA of various inversion strategies. For L1 error, pixel values are between [0, 255].}
  \label{tab:inverting}
\end{table}

	\begin{figure}[ht]
		\setlength{\tabcolsep}{0.5pt}
		\centering
    	\setlength{\fboxrule}{2pt} 
    	\setlength{\fboxsep}{0pt}
    	\scalebox{1}{
		\begin{tabular}{cccc|cccc}
			\tabcolsep0em
			\textit{\scriptsize Input} & \textit{\scriptsize -25$^{\circ}$} & \textit{\scriptsize Rec.} & \textit{\scriptsize +25$^{\circ}$} & \textit{\scriptsize Input} & \textit{\scriptsize -25$^{\circ}$} & \textit{\scriptsize Rec.} & \textit{\scriptsize +25$^{\circ}$} \\
			
	        \fbox{\includegraphics[width=0.11\linewidth]{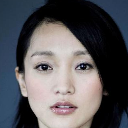}} &
			\includegraphics[width=0.11\linewidth]{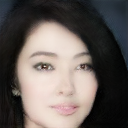} &
			\fbox{\includegraphics[width=0.11\linewidth]{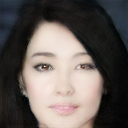}} &
			\includegraphics[width=0.11\linewidth]{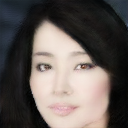} &
		    \fbox{\includegraphics[width=0.11\linewidth]{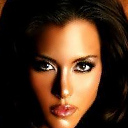}} & \includegraphics[width=0.11\linewidth]{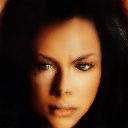} & \fbox{\includegraphics[width=0.11\linewidth]{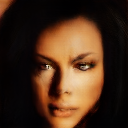}} & \includegraphics[width=0.11\linewidth]{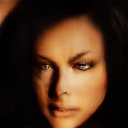}
			\\
			\fbox{\includegraphics[width=0.11\linewidth]{images/fitting/190415.png}} &
			\includegraphics[width=0.11\linewidth]{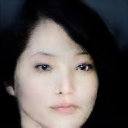} &
			\fbox{\includegraphics[width=0.11\linewidth]{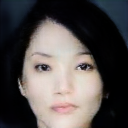}} &
			\includegraphics[width=0.11\linewidth]{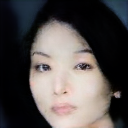} &
			\fbox{\includegraphics[width=0.11\linewidth]{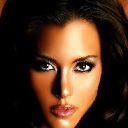}} & \includegraphics[width=0.11\linewidth]{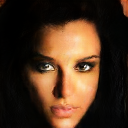} & \fbox{\includegraphics[width=0.11\linewidth]{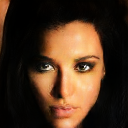}} & \includegraphics[width=0.11\linewidth]{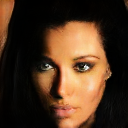} \\
			
		    \fbox{\includegraphics[width=0.11\linewidth]{images/fitting/190415.png}} &
			\includegraphics[width=0.11\linewidth]{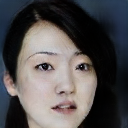} &
			\fbox{\includegraphics[width=0.11\linewidth]{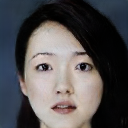}} &
			\includegraphics[width=0.11\linewidth]{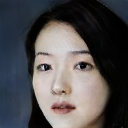} &
			\fbox{\includegraphics[width=0.11\linewidth]{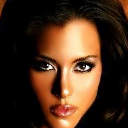}} &  \includegraphics[width=0.11\linewidth]{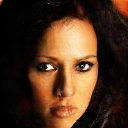} & \fbox{\includegraphics[width=0.11\linewidth]{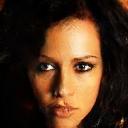}} & \includegraphics[width=0.11\linewidth]{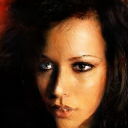} \\
			\fbox{\includegraphics[width=0.11\linewidth]{images/fitting/190415.png}} &
			\includegraphics[width=0.11\linewidth]{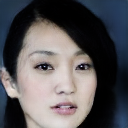} &
			\fbox{\includegraphics[width=0.11\linewidth]{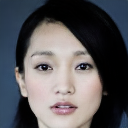}} &
			\includegraphics[width=0.11\linewidth]{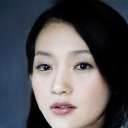}  &
			\fbox{\includegraphics[width=0.11\linewidth]{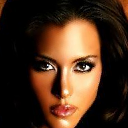}} &  \includegraphics[width=0.11\linewidth]{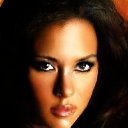} & \fbox{\includegraphics[width=0.11\linewidth]{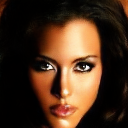}} & \includegraphics[width=0.11\linewidth]{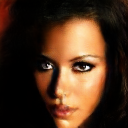}\\	
		\end{tabular}}
		\caption{Qualitative comparison between different inversion strategies.
		From top to bottom: \cite{creswellTNNL2019}, \cite{zhuECCV2016}, our Stage 2, our Stage 2 + image-specific finetuning. From left to right: input, azimuth rotations respect to the input of -25$^{\circ}$, 0$^{\circ}$(reconstruction), and 25$^{\circ}$.}
		\label{fig:fitting}
	\end{figure}

\subsection{Comparison with GAN inversion}
\label{sec:gan_inversion_exp}
We now want to explore the fidelity that our framework can achieve when inverting the generative model $D$.
In \cref{tab:inverting} we compare reconstruction results on 2000 random samples of the test set of CelebA of our Stage 2 and Stage 2 with image specific finetuning against inversion methods directly applied on $D$ at the end of stage 1.
In particular we consider \textit{fitting} the latent space of a generative model as done in ~\cite{creswellTNNL2019} or using an encoder trained to invert a frozen generator, to initialize $z$ as done in ~\cite{zhuECCV2016}. 
We focus on these two inversion methods since they are general purpose and not explicitly designed around one generator architecture, e.g., ~\cite{Abdal_2019_ICCV, Abdal_2020_CVPR} for StyleGAN. 
In \cref{fig:fitting}, we show a qualitative comparison of these approaches. 
Our Stage 2 (row 3) technique can already achieve much better performance w.r.t. the other two methods (rows 1 and 2), which demonstrates that by finetuning the weights of the generator the network achieves much higher reconstruction fidelity. Moreover, the finetuning (row 4) on a single test image dramatically improve the results, achieving impressive image reconstructions.
We notice that our method recovers fine details such as hair, eyes and skin color, and can still rotate objects meaningfully.
We also tried to alternate the fitting of z and pose to avoid ambiguities in the optimization process, achieving comparable results. These results support our intuition that by finetuning the weights of the generator (last two rows) we achieve much better reconstructions.

\begin{figure}
    \centering
    \includegraphics[width=\linewidth]{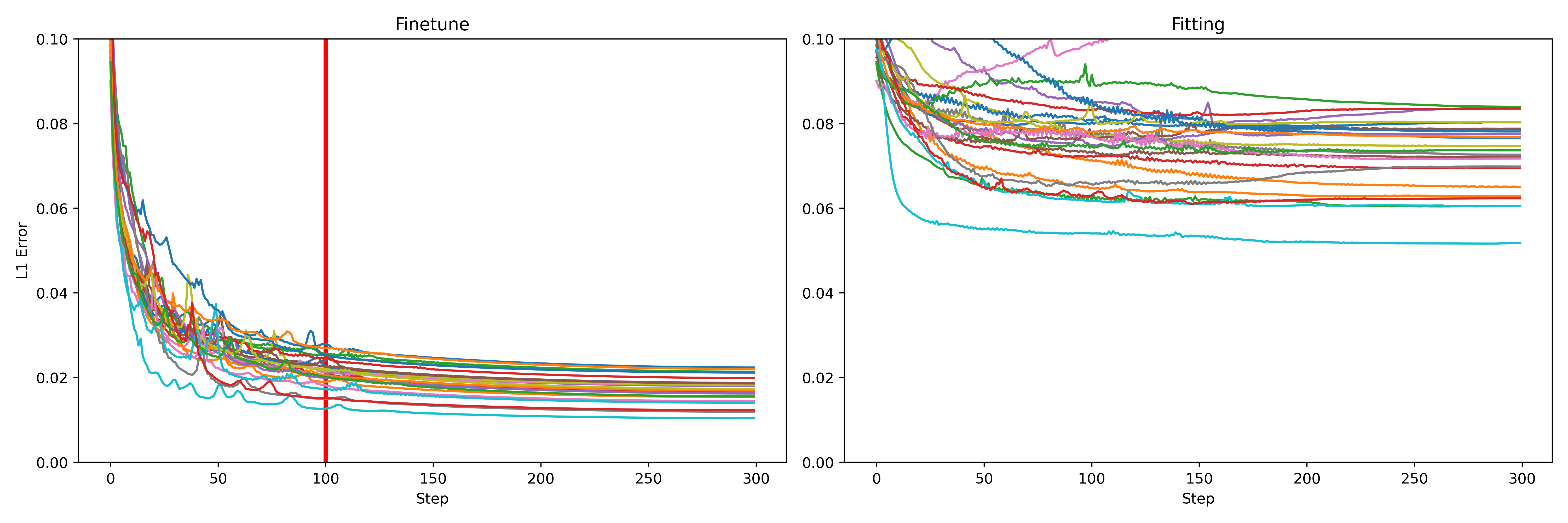} 
    \caption{L1 training reconstruction error of 25 test samples from CelebA. Left plot: \textit{Finetuning} (Ours), the red line denotes the iteration we used for early stopping in our experiments. Right plot: \textit{Fitting} with \cite{creswellTNNL2019}. Same samples have the same color in the two plots. Pixel intensities in the images are normalized in the range $[-1,1]$.}
    \label{fig:plot_fit_vs_ft}
\end{figure}

\begin{figure}[!ht]
	\centering
	\includegraphics[width=\linewidth]{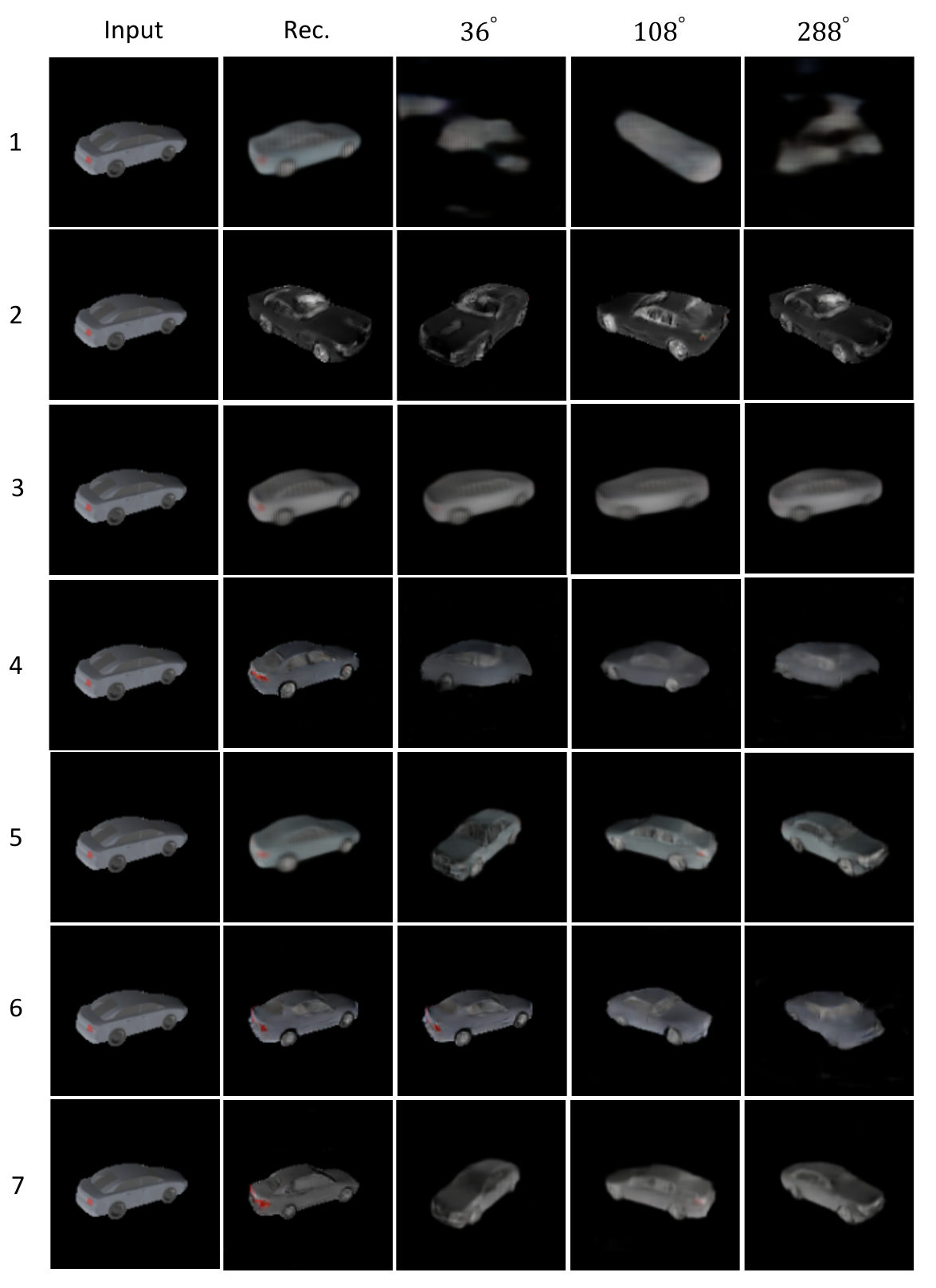} 
	\caption{Ablation qualitative results on ShapeNet Cars. The leftmost column report experiment ids corresponding to \cref{tab:ablation}.}
	\label{fig:ablation_qualitatives}
\end{figure}

In \cref{fig:plot_fit_vs_ft}, to further evaluate the effectiveness of our finetuning we plot the $L_1$ reconstruction error of our finetuning method (left plot) and \cite{creswellTNNL2019} (right plot) during training on 25 randomly sampled test images (an image per plotted line). Images are always normalized in the range $[-1,1]$ during training.
The left plot shows that our method converges to a much lower error (approximately a third of the \textit{fitting} one) and in a more stable way across test samples. 
Indeed, after only 100 optimization steps (vertical red line in left plot) we are able to perform early stopping since our method has already converged to a stable solution for almost all samples.
The right plot instead shows how for the \textit{fitting} strategy we had to train for approximately 300 steps before reaching stable solutions.

Finally, we measured the mean and the standard deviation of the time employed by our finetuning operation (100 steps) over 25 images with two different devices, a NVIDIA GTX 1080 Ti GPU and Intel i7-9700K CPU, achieving a total computation time of 3.817s $\pm$ 0.086 with the GPU and 42.661s $\pm$ 0.269 with the CPU.

\begin{table*}[ht]
\centering
\centering
\scalebox{1}{
  \begin{tabular}{l|cccccc|cc|ccc}
  \multicolumn{7}{c|}{} & \multicolumn{5}{c}{Cars} \\
  \midrule
  & & & & & & & \multicolumn{2}{c|}{NVS} & \multicolumn{3}{c}{Pose Estimation}   \\
  \toprule
    Id & \rotatebox{90}{Stage 1} & \rotatebox{90}{Autoencoder} & \rotatebox{90}{Self-Distill.} & \rotatebox{90}{Multi-view} & \rotatebox{90}{Adversarial} & \rotatebox{90}{$z$ \& $\theta$ Consist.} & $L_1\downarrow$ & SSIM$\uparrow$  & Acc.$\uparrow$ & Median$\downarrow$ & L1 Rho$\downarrow$
    \\\midrule
    1 & \xmark & \cmark & \xmark & \xmark & \xmark & \xmark & 15.38 & 0.61  & 6.85 & 91.00 & 0.018\\
    2 & \cmark & \xmark & \xmark & \xmark & \xmark & \xmark & 13.29 & 0.70 & 36.77 & 40.43 & 0.034 \\
    3 & \cmark & \cmark & \xmark & \xmark & \xmark & \xmark & 13.93 & 0.69 & 36.77 & 40.43 & 0.034 \\
    4 & \cmark & \cmark & \cmark & \xmark & \xmark & \cmark & 12.81 & 0.69 & 16.70 & 88.41 & 0.032\\
    5 & \cmark & \cmark & \cmark & \cmark & \xmark & \xmark & \textbf{7.66} & \textbf{0.77} & \textbf{86.00} & \textbf{5.73} & \textbf{0.024}\\
    6 & \cmark & \cmark & \cmark & \cmark & \cmark & \xmark &  14.36 & 0.67 &14.31 & 92.13 & 0.025 \\
    7 & \cmark & \cmark & \cmark & \cmark & \cmark & \cmark & 7.67 & 0.77 & 86.72 &  5.79 & 0.026\\
    \bottomrule
  \end{tabular}}
  \caption{Ablation study on ShapeNet Cars. For L1 error, pixel values are in [0, 255] range.}
  \label{tab:ablation}
\end{table*}

\subsection{Ablation Study}
\label{ss:two_stage_ablation}
We conduct an ablation study to show the impact of every component of our framework.
For these experiments we train our framework on ShapeNet Cars for 50 epochs in Stage 1 and 10 epochs in Stage 2, and we evaluate on the test set using L1 and SSIM (reported into the NVS columns) metrics.
We report the quantitative and qualitative NVS results in \cref{tab:ablation} and in \cref{fig:ablation_qualitatives} respectively. 
In \cref{fig:ablation_qualitatives}, from left to right, we show: input, reconstruction using the predicted $\theta$ from $E$, three azimuth rotations of 36$^{\circ}$, 108$^{\circ}$ and 288$^{\circ}$ respectively.
In row 1, we train our framework as an autoencoder with disentangled pose in the latent space and explicit rotations of deep features with a rigid geometric transformation. As shown in the picture, although the reconstruction results are good, we cannot meaningfully rotate the object due to the lack of explicit multi-view supervision
In row 2 we show the results of the autoencoder after pre-training with Stage 1. Even though the objects can rotate correctly, the car appearance is different and the pose in the reconstruction is wrong.
In row 3 we add the Stage 1 pre-training before rebuilding the autoencoder.
The model cannot rotate the object, however, each pose generates a realistic car even if with the wrong orientation. We argue that during the second stage the model forgets the good 3D representation learnt during the first stage, supporting our claim.
To fight forgetting, we add self-distillation (row 4) and try to reconstruct the generated samples from $D_f$ including also the consistency losses $L_z$ and $L_{_\theta}$.
With the former strategy, we can generate novel views of the object, but they are not consistent.
Thus, to improve 3D representations, we add the multi-view supervision (row 5) on the generated samples which can be obtained for free.
After this addition we are able to rotate the object showing that this is the crucial step for learning 3D features.
To further improve the quality of the generated sample, we add an adversarial loss similar to ~\cite{Isola_2017_CVPR} on the real samples (row 6). However, the instability of the adversarial training makes the training prone to collapsing. Finally, with the consistency losses (row 7) $L_z$ and $L_{_\theta}$ we can stabilize the adversarial training. 
From our experiments, the adversarial loss is fundamental in real datasets to achieve sharp results, therefore we kept it on our formulation, even if it does not improve on synthetic datasets.

\subsection{Unsupervised Pose Estimation}
We report here results regarding the poses inferred by our method from the input image.
When evaluating, given the estimated and ground truth poses on the training and validation sets, we first train a linear regressor to map poses from the reference frame implicitly learnt by our model to the ground-truth one of ShapeNet. Then we use it at test time to compare predicted poses to ground truth ones.
We note that the scale parameter used in $D$ and the distance to the center of the world $\rho$ used in ShapeNet are inversely linearly related. 
In terms of metrics, we compute the absolute $\rho$ error ($\rho$ ranging in $[1.0,1.5]$) and the angle error as the cosine distance between the two azimuths (the angle with 360$^\circ$ range) and report the median for both. 
Moreover, we calculate the angle prediction accuracy, defined as the ratio of samples with an angle error smaller than 30$^\circ$ with the total number of samples \cite{Tulsiani_2018_CVPR}.

In the last 3 columns of \cref{tab:ablation}, we report quantitative pose estimation performance for the Car class of ShapeNet for each combination of the ablation study.
The table highlights once more how the key ingredients of our formulation are the self-distillation and multi-view training used in Stage 2.
In particular the multi-view training on generated images (row 5), feasible only thanks to our self-distillation, doubles the performance compared to the network after Stage 1 (row 2). Moreover, we notice that the network fails when trained without distillation (row 3), failing to generate objects that rotate coherently.

\begin{figure}[ht]
	\centering
	\includegraphics[width=0.95\linewidth]{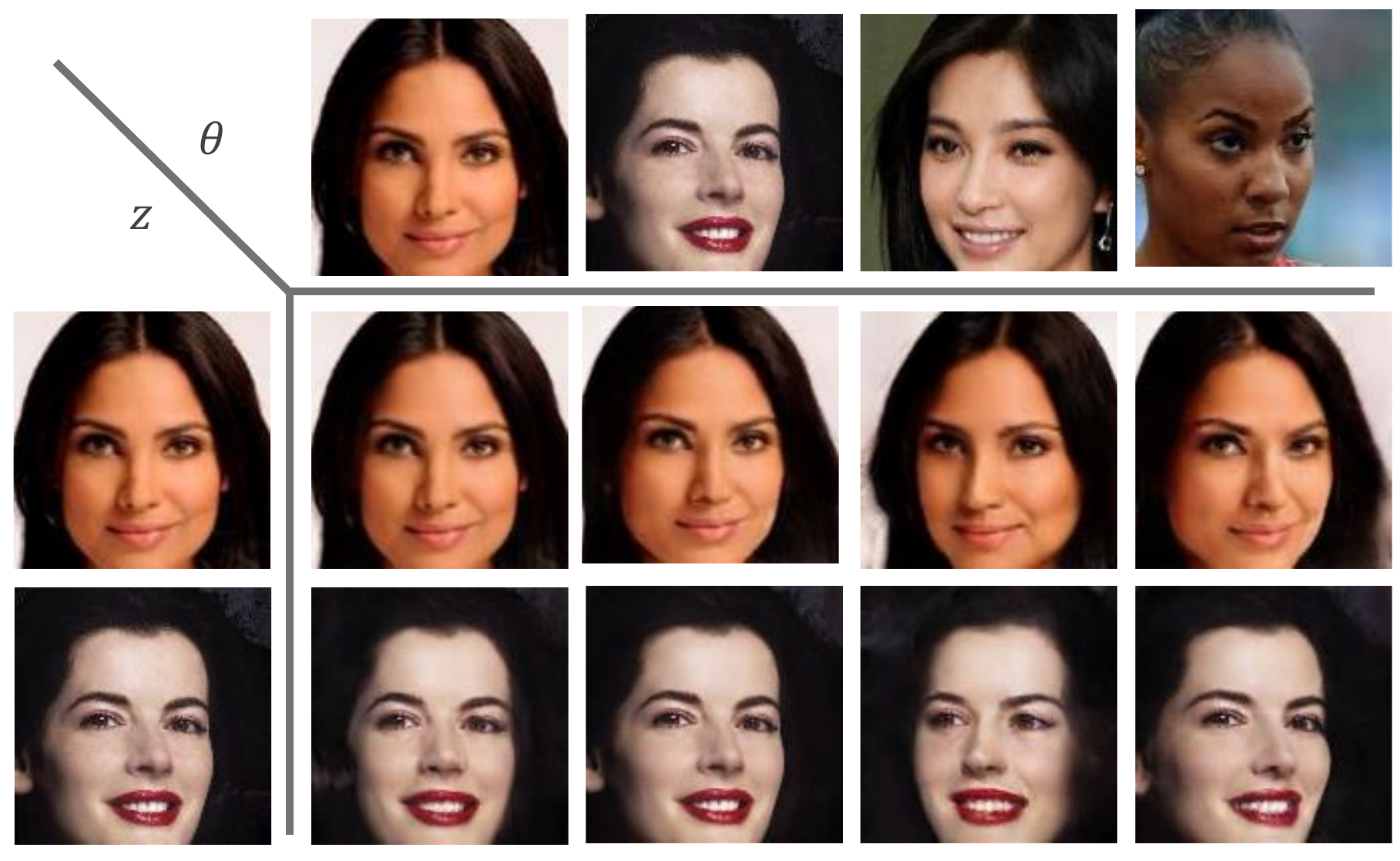} 
	\caption{Qualitative pose estimation on CelebA. 
	We extract $\theta$ from $I_{\theta}$ (row 1) and $z$ from $I_z$ (column 1) to craft a novel image.}
	\label{fig:swap_pose_celebA}
\end{figure}

On real datasets we do not estimate the pose error quantitatively due to the lack of GT, therefore we provide a qualitative evaluation in \cref{fig:swap_pose_celebA}.
Given two images $I_z$ and $I_{\theta}$, we infer the latent representation $z$ from $I_z$ and the pose $\theta$ from $I_{\theta}$. 
Then, we can generate a new image with the appearance of $I_z$ and the pose of $I_{\theta}$ by simply combining the estimated $z$ and $\theta$. 
\cref{fig:swap_pose_celebA} shows the results of this experiment on the test set of CelebA. We employ the network after finetuning for this experiment.
Notably, our method can estimate with high accuracy also challenging poses, such as faces seen from the side, which is an underrepresented pose in the training set.

\begin{figure}[ht]
    \newlength{\grafmultiplestagesWidth}
    \setlength{\grafmultiplestagesWidth}{0.18\linewidth}
    \centering
    \begin{tabular}{c|ccc}
        \toprule
        Input & Stage 1 & Stage 2 & Finetune \\
        \midrule
        \includegraphics[width=\grafmultiplestagesWidth]{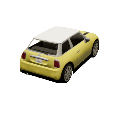} &
        \includegraphics[width=\grafmultiplestagesWidth]{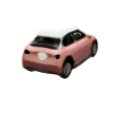} &
        \includegraphics[width=\grafmultiplestagesWidth]{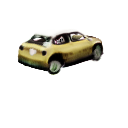} &
        \includegraphics[width=\grafmultiplestagesWidth]{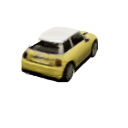} \\
        \includegraphics[width=\grafmultiplestagesWidth]{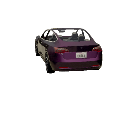} &
        \includegraphics[width=\grafmultiplestagesWidth]{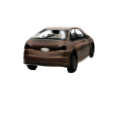} &
        \includegraphics[width=\grafmultiplestagesWidth]{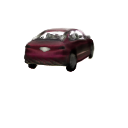} &
        \includegraphics[width=\grafmultiplestagesWidth]{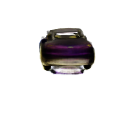} \\
        \includegraphics[width=\grafmultiplestagesWidth]{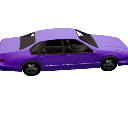} &
        \includegraphics[width=\grafmultiplestagesWidth]{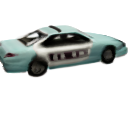} &
        \includegraphics[width=\grafmultiplestagesWidth]{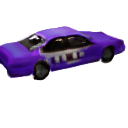} &
        \includegraphics[width=\grafmultiplestagesWidth]{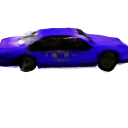} \\
    \bottomrule
    \end{tabular}
    \caption{Qualitative comparison of a Stage 1, Stage 2 and finetuning network using as decoder GRAF ~\cite{schwarz2020graf} instead of HoloGAN ~\cite{Nguyen-Phuoc_2019_ICCV} on the Carla dataset.}
    \label{fig:graf_multiple_stages}
\end{figure}

\begin{figure}[ht]
    \newlength{\grafRotationWidth}
    \setlength{\grafRotationWidth}{0.115\linewidth}
    \centering
    \begin{tabular}{c|ccccc}
        \toprule
        Input & \multicolumn{5}{c}{$\theta$ Interpolation}\\
        \midrule
        \includegraphics[width=\grafRotationWidth]{images/graf/008000/input.png} &
        \includegraphics[width=\grafRotationWidth]{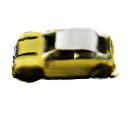} &
        \includegraphics[width=\grafRotationWidth]{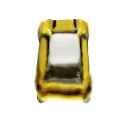} &
        \includegraphics[width=\grafRotationWidth]{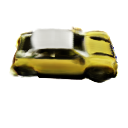} &
        \includegraphics[width=\grafRotationWidth]{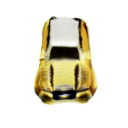} &
        \includegraphics[width=\grafRotationWidth]{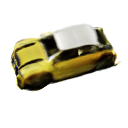} \\
        
        \includegraphics[width=\grafRotationWidth]{images/graf/008005/input.png} &
        \includegraphics[width=\grafRotationWidth]{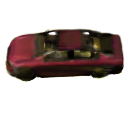} &
        \includegraphics[width=\grafRotationWidth]{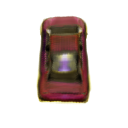} &
        \includegraphics[width=\grafRotationWidth]{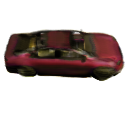} &
        \includegraphics[width=\grafRotationWidth]{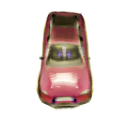} &
        \includegraphics[width=\grafRotationWidth]{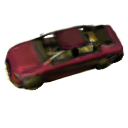} \\
        
        \includegraphics[width=\grafRotationWidth]{images/graf/008007/input.png} &
        \includegraphics[width=\grafRotationWidth]{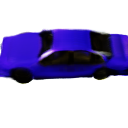} &
        \includegraphics[width=\grafRotationWidth]{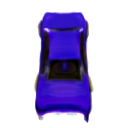} &
        \includegraphics[width=\grafRotationWidth]{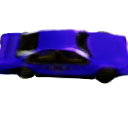} &
        \includegraphics[width=\grafRotationWidth]{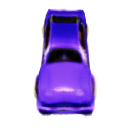} &
        \includegraphics[width=\grafRotationWidth]{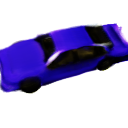}\\
        \bottomrule
    \end{tabular}
    \caption{Novel view synthesis from a single input image using as our decoder GRAF ~\cite{schwarz2020graf} instead of HoloGAN ~\cite{Nguyen-Phuoc_2019_ICCV} on the Carla dataset.}
    \label{fig:graf_rotations}
\end{figure}

\subsection{Ablation on learned perceptual metrics}
\label{sec:adversarial}

\begin{figure}
	\setlength{\tabcolsep}{1pt}
	\centering
	\setlength{\fboxrule}{2pt} 
	\setlength{\fboxsep}{0pt} 
	\scalebox{0.9}{
	\begin{tabular}{cc|cccc}
	&Input. & \multicolumn{4}{|c}{$\theta$ Interpolation} \\
	\rotatebox{90}{\hspace{5mm}Full} & 
	 \includegraphics[width=0.095\textwidth]{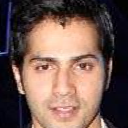} & \includegraphics[width=0.095\textwidth]{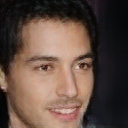} & \includegraphics[width=0.095\textwidth]{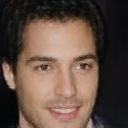} & \includegraphics[width=0.095\textwidth]{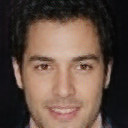} & \includegraphics[width=0.095\textwidth]{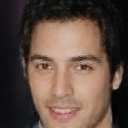} \\

    \rotatebox{90}{\hspace{3mm} Pixel} 
& \includegraphics[width=0.095\textwidth]{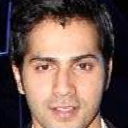} & \includegraphics[width=0.095\textwidth]{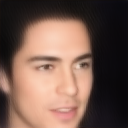} & \includegraphics[width=0.095\textwidth]{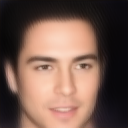} & \includegraphics[width=0.095\textwidth]{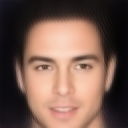} & \includegraphics[width=0.095\textwidth]{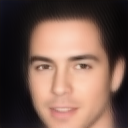} \\

    \rotatebox{90}{\hspace{5mm}Full} & 
\includegraphics[width=0.095\textwidth]{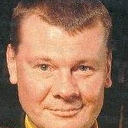} & \includegraphics[width=0.095\textwidth]{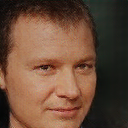} & \includegraphics[width=0.095\textwidth]{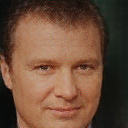} & \includegraphics[width=0.095\textwidth]{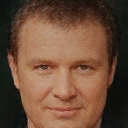} & \includegraphics[width=0.095\textwidth]{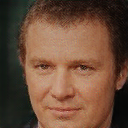} \\
    
    \rotatebox{90}{\hspace{3mm} Pixel} &  
\includegraphics[width=0.095\textwidth]{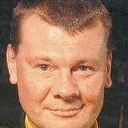} & \includegraphics[width=0.095\textwidth]{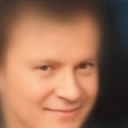} & \includegraphics[width=0.095\textwidth]{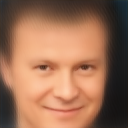} & \includegraphics[width=0.095\textwidth]{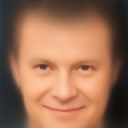} & \includegraphics[width=0.095\textwidth]{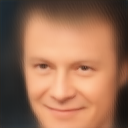} \\
	\end{tabular}}
	\caption{Qualitative comparison between results of Stage 2 with or without losses computed on deep feature spaces.}
	\label{fig:adv_ablation}
\end{figure}

In our stage 2 we use multiple losses to measure the similarity between reconstructed images and real ones. 
Using a very high level subdivision they can be split into losses operating directly on pixel intensities ($L_2$ and $L_{SSIM}$) and losses operating on a learned deep feature space ($L_{VGG}$ and $Adversarial$).
To showcase the impact of the latter for the quality of the generated images we designed an ad-hoc comparison by training a variant of our model where in stage 2 we optimize only pixel intensities losses (\ie, $L_2$ and $L_{SSIM}$).
\cref{fig:adv_ablation} collects some qualitative results comparing this version of the model to our full formulation.
The images generated by the network trained with all losses have two clear advantages with respect to the alternative: they are sharper and contain more details specific of the provided input image (\eg the beard of the man in the top or the hair of the man in the bottom). 
Both models are able to generate images that resemble the input image.

\subsection{Preliminary Results with a Different Decoder}
\label{sec:graf}

We report preliminary results obtained changing the decoder network in our formulation with the recently proposed neural radiance field based GRAF ~\cite{schwarz2020graf}.
While the architecture of the network is quite different from the HoloGAN ~\cite{Nguyen-Phuoc_2019_ICCV} one used throughout the paper, from a black box perspective, both models allow to provide a latent vector and a pose to generate a view of an object and can therefore be in principle interchanged in our framework.

While the swap does not require major changes, the computational cost of rendering a full image using GRAF is significantly higher.
Thus, following the same strategy as its authors, we train on patches.
However, the encoder requires an entire image to estimate $z$ and $\theta$. Thus, in stage 1, we initialize GRAF with the pretrained weights released by the authors, and we train only the encoder $E$ to invert images to the generator latent space.
Then, in stage 2 due to the patch generation we did not use $L_{VGG}$ and $L_{SSIM}$ to measure the reconstruction error, but we rely only on the pixel-wise $L_2$ loss. . 

In \cref{fig:graf_multiple_stages}, we report results obtained from a Stage 1, Stage 2 and a Fine-tuned network on the Carla dataset ~\cite{schwarz2020graf}. As we can see from the picture, Stage 1  is not sufficient to obtain a good reconstruction, while we can recover the identity of the input car in Stage 2. Then, during the finetuning we recover some details of the input image such as the color and the pose. In \cref{fig:graf_rotations} we report some qualitative results of generated novel views for the same input images showing how the multi-view consistency of the models is maintained.

\section{Method Limitations}
\label{sec:limitations}
We notice that sometimes $D$ during Stage 1 fails to converge for some datasets. In such cases, we inherit the same artifacts during Stage 2.
For instance, we tried to train on the Bench and on the Chair classes of ShapeNet, however the HoloGAN based $D$ was not able converge, completely collapsing or learning incomplete representations that allow only partial rotations of the objects. In such case, our Stage 2  learns the same incomplete rotations while distilling the generative knowledge. 
Moreover, we notice that HoloGAN is performing much better on real rather than synthetic datasets. 
We think that due to the comparatively larger amount of samples in the real datasets the unsupervised learning strategy works better.

\section{Conclusions}
In this paper we have presented a novel method for NVS from a single view. 
Our method is the first of its kind that can be trained without requiring any form of annotation or pairing in the training set, the only requirement being a collection of images of objects belonging to the same macrocategory.
Our method achieves performance comparable to other alternatives on synthetic datasets, while also working on real datasets like CelebA or Cars, where no ground truth is available and the competitors cannot be trained.
We tested our framework on NVS with only a single object in the scene. In future, we plan to extend our experiments to scenes containing multiple objects leveraging recently proposed models for scene generation ~\cite{nguyenphuoc2020blockgan,niemeyer2021giraffe}.

{\small
\bibliographystyle{ieee_fullname}
\bibliography{egbib}
}
\end{document}